\documentclass{article}

\PassOptionsToPackage{numbers,sort&compress}{natbib}
\usepackage[main, final]{neurips_2026}

\usepackage[utf8]{inputenc} % allow utf-8 input
\usepackage[T1]{fontenc}    % use 8-bit T1 fonts
\usepackage{hyperref}       % hyperlinks
\usepackage{url}            % simple URL typesetting
\usepackage{booktabs}       % professional-quality tables
\usepackage{amsfonts}       % blackboard math symbols
\usepackage{nicefrac}       % compact symbols for 1/2, etc.
\usepackage{microtype}      % microtypography
\usepackage{xcolor}         % colors

\usepackage{algorithm}
\usepackage{algorithmic}
\usepackage{multirow}
\usepackage{tabularx}
\usepackage{amsfonts,bm}
\usepackage{threeparttable}
\usepackage{upgreek}
\usepackage{pifont}%
\usepackage{bbding}
\usepackage{makecell}
\usepackage{color}
\usepackage{xcolor}
\usepackage{colortbl}
\usepackage{wrapfig}
\usepackage{soul}
\usepackage[nointegrals]{wasysym}
\usepackage{xspace}
\usepackage{graphicx}
\usepackage{subcaption}
\usepackage[first=0,last=9]{lcg}
\usepackage{stackengine}
\usepackage{amsmath}
\usepackage{amssymb}
\usepackage{wrapfig}

\definecolor{Gray}{gray}{0.85}
  
\newcolumntype{x}[1]{>{\centering\arraybackslash}p{#1pt}}
\newcolumntype{y}[1]{>{\raggedright\arraybackslash}p{#1pt}}
\newcolumntype{z}[1]{>{\raggedleft\arraybackslash}p{#1pt}}

\definecolor{neuripspink}{RGB}{219, 112, 147}
\newcommand{\mypinklink}[2]{%
  \begingroup
    \hypersetup{pdfborder={0 0 0}, colorlinks=false}% Removes the box locally
    \href{#1}{\color{neuripspink}{#2}}% Applies the pink text color
  \endgroup
}

\definecolor{sectiongray}{RGB}{242,242,242}
\definecolor{methodblue}{RGB}{226,233,246}
\definecolor{methodbg}{RGB}{238,241,247}   
\definecolor{deltagreen}{RGB}{220,235,220}
\definecolor{goodgreen}{RGB}{90,200,90}
\title{Back2Struct: Making Structured Images Editable Again}

\author{%
  Pengyu Yan \qquad
  Yixin Wu \qquad
  Yunjie Tian \qquad
  David Doermann \\[2mm]
  University at Buffalo, SUNY \\
  \texttt{\{pyan4,yixinwu,yunjieti,doermann\}@buffalo.edu}
}

\begin{document}

\newcommand{\model}{Back2Struct\xspace}

\maketitle

\begin{center}
\includegraphics[width=1.0\linewidth,height=0.65\linewidth]{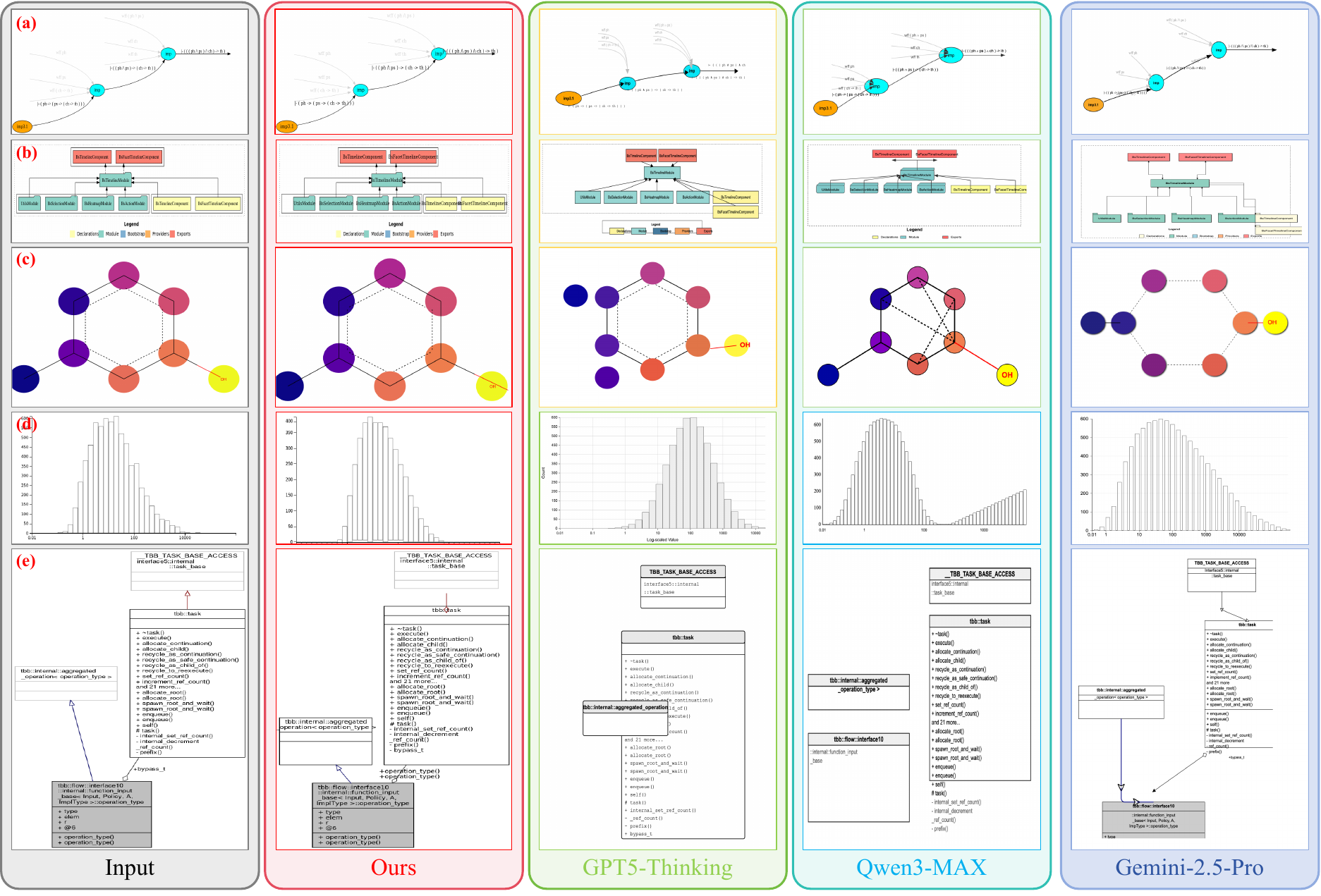}
\captionof{figure}{\textbf{A qualitative comparison on recovering the reference images among the most advanced LLMs,} including GPT-5 (Thinking mode)~\cite{singh2025openai_gpt5}, Qwen3-Max (over 1 trillion parameters)~\cite{yang2025qwen3}, Gemini-2.5-Pro~\cite{comanici2025gemini25}, and \model{} (Ours). \model{} exceeds its counterparts with negligible confusion or mistakes in its responses, showcasing its superior ability in ``making structured images editable again''. Note that \model{} has only 7B parameters.}
\label{fig:first_show}
\end{center}

\begin{abstract}
Structured images, such as diagrams, charts, and flowcharts, are inherently symbolic and can be compactly represented in an editable format, yet in practice, they are often rendered as images, and therefore not graphically editable.
This mismatch presents a significant challenge for researchers, engineers, and designers who wish to incorporate modified versions of existing graphic content into new materials without manually reconstructing it.
In this study, we present \textbf{\model}, which ``makes structured images editable again'' by directly recovering vector graphics code (SVG / XML) from image representations.
Given an image of a structured graphic, \model predicts semantically object-level SVG / XML code that explicitly encodes text, shapes, topology, and layout, rather than performing low-level pixel vectorization.
The generated code can be seamlessly imported into tools such as PowerPoint, allowing users to edit, refine, restyle, and reuse graphic content while preserving structural fidelity.
Beyond supervised fine-tuning on ground-truth SVG token sequences, we further optimize \model with reward-based learning to better match deployment-time requirements: the output should be syntactically valid, properly concise, and visually faithful to the input diagram.
Specifically, we design a composite reward that jointly encourages SVG / XML compilability, length consistency with the reference code, and structural or semantic similarity between the generated and ground-truth graphics.
These complementary signals guide the model to produce SVGs that are not only closer to the training distribution, but also more complete, editable, and renderable in practice.
Experiments show that \model improves accuracy, editability, validity, and user alignment over baselines.
Dataset and code are available at: \mypinklink{pengyu965.github.io/Back2Struct.github.io}{pengyu965.github.io/Back2Struct.github.io}
\end{abstract}

%%%%%%%%%%%%%%%%%%%%%%%%%%%%%%%%%%%%%%%%%%%%%%%%%%%%%%%%%%%%%%%%%%%%%%%%%%%%%%%%%%%%%%%%%%%%%%%%%%%%%%%%%%%%%%%%%%%%%%%%%%%%%%%%%%%%%%%%%%%%%

\section{Introduction}
\label{sec:intro}

%-------------------------------------------------------------------------
Structured content, including diagrams, charts, flowcharts, user interface designs, and block architectures, is central to how researchers, engineers, and designers communicate algorithms, pipelines, system architectures, and design intent.
Unlike natural images, its utility depends on discrete symbolic structures, such as nodes and edges, groupings and alignments, textual labels, geometric primitives, and layout regularities that encode precise relationships rather than appearance alone.
These structures can in principle be compactly represented in editable formats such as SVG / XML~\cite{rodriguez2025starvector, svg, li2020differentiable, ma2022towards, hahn2014autotrace}, making them interpretable, verifiable, and directly modifiable.
However, structured graphics are often shared as screenshots, raster exports, or figures embedded in papers and slides.
Once rendered into pixels, their symbolic structure becomes inaccessible: individual elements cannot be selected, text and shapes cannot be reliably edited, and constraints such as alignment, grouping, and connectivity are lost.
As a result, even simple edits may require manually reconstructing the entire graphic, motivating the recovery of an editable and semantically meaningful representation.

Existing approaches provide only partial solutions.
Pixel-level generative models can synthesize visually plausible images, but they struggle to enforce global structural constraints and produce non-editable bitmaps~\cite{goodfellow2020gan, ho2020ddpm, nichol2021improvedDDPM, rombach2022highLDM}.
Conventional vectorization methods recover curves, contours, and low-level shapes, but not high-level semantics such as node-edge relationships, hierarchical groupings, textual associations, or layout intent~\cite{belouadi2023automatikz, belouadi2024detikzify, rodriguez2023figgen}.
Recent multimodal large language models~\cite{alayrac2022flamingo, achiam2023gpt4, team2023gemini} and their successors~\cite{liu2023llava, bai2025qwen25vl} show strong visual understanding and code-generation abilities, yet they are not explicitly optimized for recovering faithful, executable, and practically usable SVG / XML code from realistic structured images~\cite{rodriguez2025starvector, song2023clipvg, lin2024live, hahn2014autotrace}.
In particular, supervised fine-tuning on ground-truth SVG token sequences mainly encourages code imitation, but does not directly optimize deployment-time properties such as validity, conciseness, and visual faithfulness.

In this work, we argue that making structured images editable again requires treating them as symbolic artifacts rather than ordinary raster images.
We introduce \textbf{\model}, a framework that directly recovers manipulatable vector graphics code from image representations, as shown in Figure~\ref{fig:first_show}.
Given an image of structured content, \model predicts object-level SVG / XML code that explicitly encodes text, shapes, topology, and layout, rather than performing low-level pixel vectorization.
The recovered code can be imported into standard authoring tools such as PowerPoint, allowing users to edit, refine, restyle, and reuse graphic content while preserving structural fidelity and interpretability.
To better match practical editing needs, we further optimize \model beyond supervised fine-tuning with reward-based learning.
We design a composite reward that captures complementary requirements: SVG / XML compilability, length consistency with the reference code, and structural or semantic similarity between the generated and ground-truth graphics.

These reward signals encourage outputs that are syntactically valid, neither truncated nor unnecessarily verbose, and faithful to the original visual structure and semantic content.
As a result, the model generates SVGs that are not only closer to the training distribution, but also more complete, editable, and usable in real workflows.
We evaluate \model on representative structured image recovery scenarios where both semantic correctness and layout fidelity are critical.
Experiments show that \model improves structural accuracy, editability, SVG validity, and alignment with user intent over strong vision--language and code-generation baselines.
Overall, the results suggest that code-centric recovery, combined with deployment-aware reward optimization, is an effective way to turn static structured images back into editable visual artifacts.

The main contributions of this paper are as follows.
\begin{itemize}
\item We collect and curate the \textbf{StructHub} dataset containing \textbf{84K} paired structured raster images and their associated SVG/XML code, together with an evaluation benchmark. This is a large-scale, \textbf{high-value} dataset dedicated to editable structured image generation.

\item We propose \textbf{\model}, a framework that makes structured images editable again by recovering semantically meaningful SVG / XML code from image representations, and formulate this task as object-level vector graphics code generation that explicitly models text, shapes, topology, and layout.

\item We introduce reward-based optimization to improve deployment-time properties of the generated SVG / XML code, including validity, length consistency, and structural or semantic faithfulness. Extensive experiments show that \model improves structural accuracy, editability, SVG/XML validity, and alignment with user intent compared with strong multimodal and code-generation baselines.

\end{itemize}

\section{Related Work}
\label{sec:related_work}

\noindent\textbf{Multi-modal Large Language Models.}
Recent large vision--language models (LVLMs) extend large language models (LLMs) from text-only reasoning to visual inputs by jointly modeling images and text.
Representative models, including LLaVA~\cite{liu2023llava}, InternVL~\cite{chen2024internvl}, InstructBLIP~\cite{dai2023instructblip}, Qwen-VL~\cite{qwen2,qwen2.5,qwen3}, Gemini~\cite{team2023gemini,comanici2025gemini25,team2024gemini}, and GPT~\cite{brown2020language,achiam2023gpt4,singh2025openai_gpt5}, enable open-ended visual reasoning, captioning, and grounding across diverse inputs such as documents, charts, diagrams, and UI screenshots.
However, they typically rely on continuous visual features and produce free-form text, which limits their ability to recover fine-grained symbolic structures or reconstruct compositional visual inputs in a faithful and executable form.
This motivates our study of structured image understanding, where visual perception is translated directly into editable graphics.

\noindent\textbf{Structured Image Understanding.}
Vision--language pretraining has improved image understanding by aligning visual perception with natural language and capturing high-level semantics beyond low-level pixels~\cite{vinyals2015show,xu2015show,wang2022end,radford2021learning,li2022blip,li2021align}.
Many works further study structured visual domains, including graph understanding~\cite{xing2025unifyingenhancinggraphtransformers,jain2024integrating,zala2023diagrammergpt,pan2024flowlearn}, table understanding~\cite{zhong2020pubtabnet,smock2022pubtables,nassar2022tableformer,yin2020tabert,zhang2024tablellm}, chart understanding~\cite{luo2021chartocr,liu2023deplot,liu2023matcha,yan2023context,ahmed2023spaden,yan2024chartreformer,han2023chartllama,xu2024chartmoe}, and document or layout understanding~\cite{lee2023pix2struct,huang2022layoutlmv3,kim2022donut}.
These methods mainly treat structured images as information sources for question answering, field extraction, or semantic interpretation.
In contrast, our goal is to faithfully reconstruct the original structured image by recovering its content, topology, layout, and style as editable vector graphics.

\noindent\textbf{Image-to-Code Generation with LVLMs.}
Image-to-code generation converts visual inputs into executable or markup representations, providing a natural path toward editable visual understanding.
Early works such as Im2LaTeX~\cite{deng2017image} and pix2code~\cite{beltramelli2018pix2code} translate mathematical expressions or GUI screenshots into LaTeX, HTML, DOM, or DSL tokens~\cite{chen2020ui2code,lee2023pix2struct}.
In vector and diagram domains, methods such as DiffVG~\cite{li2020diffvg} and DeepSVG~\cite{carlier2020deepsvg} generate SVG-style representations through differentiable rendering or tokenized path decoding.
Recent multimodal and LLM-based methods further recover plotting code, graph specifications, or flow logic from charts, diagrams, and flowcharts~\cite{liu2023deplot,xu2024chartmoe,han2023chartllama,yan2024chartreformer,zala2023diagrammergpt,pan2024flowlearn,rodriguez2025starvector}.

Our work focuses on reconstructing diagrams as editable SVGs.
Instead of relying on high-level plotting languages such as Python, which may introduce ambiguity and lose visual fidelity, we adopt object-level SVGs to couple fine-grained image understanding with structured output generation.

\begin{figure*}[t]
\centering
\includegraphics[width=0.99\linewidth]{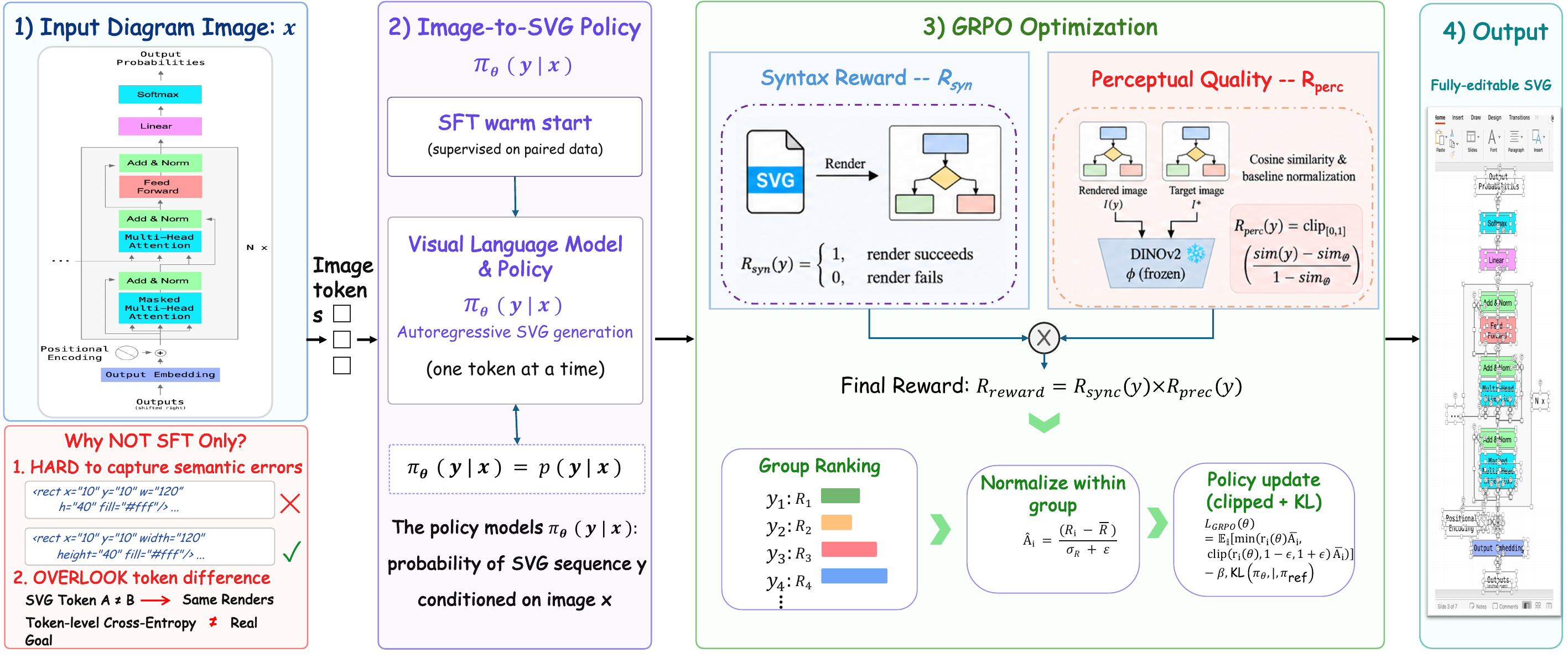}
\vspace{-4pt}
\caption{\textbf{Overview of \model}. Given an RGB reference image and/or text, \model generates object-level XML code that can be losslessly converted into an editable PowerPoint file.} 
\label{fig:method}
\vspace{-6pt}
\end{figure*}

\section{Method}

\subsection{Motivation}

Recovering SVG code from a diagram image can be formulated as a \textit{sequence generation} problem, where an input image $x$ is mapped to an XML token sequence $y$ forming a valid SVG document.
However, token-level correctness is insufficient: the output should be syntactically valid, structurally faithful, and perceptually consistent with the input.
Standard supervised fine-tuning with cross-entropy loss provides only a token-level surrogate and is poorly aligned with these goals.
For example, a small syntax error may invalidate the entire SVG, while visually equivalent SVGs can have different token sequences due to alternative attribute orders or coordinate forms.
This \textit{objective mismatch} motivates holistic reward signals that directly evaluate validity, structure, and visual quality, despite being often non-differentiable and difficult to optimize with standard supervised learning.

We therefore frame image-to-SVG generation as a reinforcement learning problem. Let $\pi_\theta(y \mid x)$ denote the policy that generates an SVG sequence $y$ conditioned on an input diagram image $x$. Our goal is to maximize the expected task reward:
\begin{equation}
    \mathcal{J}(\theta) = \mathbb{E}_{x \sim \mathcal{D},\; y \sim \pi_\theta(\cdot \mid x)}
    \bigl[R(y, y^*)\bigr],
\end{equation}
where $y^*$ denotes the ground-truth SVG and $R$ is a composite reward that measures syntactic validity, structural correspondence, and perceptual quality. The framework is illustrated in Figure~\ref{fig:method}.

\subsection{Reward Design}

Re-producing high-quality SVGs demands two separate assurances: the result must be
a \emph{usable} artifact—one that actually renders into an image which is viewable
 and editable—and it must be \emph{visually faithful} to the intended
diagram. These criteria are interdependent: an SVG can be syntactically valid yet depict the wrong
diagram, while a visually convincing attempt might not render at all. We formalize
these with a rendering \emph{gate} and a perceptual \emph{quality} term, multiplied
together so that successful rendering is treated as a prerequisite for—rather than
a part of—the reward.

\vspace{0.5em}\noindent\textbf{Syntactic Reward.}
Let $\texttt{syntactic}(y)$ obtain the SVG code from a generation $y$ and rasterize
it onto a white canvas at the target image resolution. The syntactic reward is
\begin{equation}
    R_{\text{syntactic}}(y) = \mathbf{1}\bigl[\texttt{render}(y)\ \text{succeeds}\bigr]
    \in \{0, 1\}.
\end{equation}
This enforces a stricter validity condition than mere XML well-formedness: successful
rendering implies the output is both parseable and rasterizable, thereby also filtering
out outputs that are syntactically valid yet cannot be rendered. If a generation fails
to render, it is assigned zero overall reward, since it cannot be displayed or edited.

\vspace{0.5em}\noindent\textbf{Perceptual Reward.}
We assess diagram-level quality directly in pixel space, not over SVG tokens, so
that generations are incentivized based on what they \emph{look like} rather
than how the underlying code is formatted. Let $I(y)$ denote the rendered
prediction and let $I^*$ be the target image the model is conditioned on. A
frozen DINOv2 ViT-B/14 encoder $\phi$ maps each image to an $\ell_2$-normalized
embedding; we compute the cosine similarity between embeddings and clamp it to
$[0,1]$,
\begin{equation}
    \mathrm{sim}(y) =
    \operatorname{clip}_{[0,1]}\cos\!\bigl(\phi(I(y)),\, \phi(I^*)\bigr).
\end{equation}
However, the raw similarity is a biased proxy for quality: even an
\emph{content-free} output can achieve nonzero similarity to the target. In
particular, a blank canvas $I_\varnothing$ with matching dimensions yields a
nontrivial baseline
$\mathrm{sim}_\varnothing = \operatorname{clip}_{[0,1]}\cos(\phi(I_\varnothing),\,
\phi(I^*))$, because diagrams in our setting are mostly white. To account for
this, we reward fidelity \emph{relative to} the content-free baseline by
subtracting the per-image floor and then renormalizing:
\begin{equation}
    R_{\text{perc}}(y) =
    \operatorname{clip}_{[0,1]}\!\left(
    \frac{\mathrm{sim}(y) - \mathrm{sim}_\varnothing}{1 - \mathrm{sim}_\varnothing}
    \right).
    \label{eq:perc}
\end{equation}
With this definition, $R_{\text{perc}}$ gives $0$ to a blank (or
blank-equivalent) render and $1$ to a perceptually accurate one, with
calibration performed separately for each target so the reward captures only
the fidelity gained by actually drawing meaningful content.

\vspace{0.5em}\noindent\textbf{Gated Composition.}
Our composition also removes the need for a separate length-fidelity reward. We
define the final reward as the product of the render gate and the perceptual term:
\begin{equation}
    R(y) = R_{\text{syntactic}}(y) \cdot R_{\text{perc}}(y) \in [0, 1].
    \label{eq:reward}
\end{equation}
Gating, instead of a weighted sum, is deliberate: an SVG cannot be rendered, there is no
resulting image to evaluate, so the rendering term functions as a $\{0,1\}$ mask rather than a
 separately optimizable reward component. We settled on this design after observing that an additive
form, $R = w_r R_{\text{syntactic}} + w_l R_{\text{length\_fidelity}} + w_p R_{\text{syntactic}}$, is reward-hackable: the policy collapses onto short, trivially
renderable but almost empty SVGs, because (i) successful rendering gives a constant bonus
 available to any syntactically valid output, and (ii) without baseline subtraction, an empty image can still achieve a high score when compared to the target image, which contains only sparse pixel information (Appendix Fig.~\ref{fig:imglvl_evalissue}). Baseline subtraction
(Eq.~\ref{eq:perc}) addresses (ii), and multiplicative gating (Eq.~\ref{eq:reward})
eliminates (i), ensuring the reward increases only when the output both renders \emph{and}
adds visually accurate content. Together, these mechanisms also replace the needs of 
explicit length-fidelity term: under-generation (blank or
sparse images) is pushed toward zero by the baseline, while runaway
over-generation that fails to terminate is suppressed by the gate. This rule-based reward design with a gating mechanism achieves the best performance, while avoiding the need for a complex reward design or an LLM-based reward judge (Section~\ref{sec:abla})).

\subsection{Policy Optimization with GRPO}

We optimize the SVG generation policy $\pi_\theta$ with Group Relative Policy
Optimization (GRPO)~\cite{shao2024deepseekmath_grpo}, which forgoes a learned
value network and instead estimates advantages by comparing completions sampled
for the same input. This suits image-to-SVG generation, where absolute reward
scales vary widely across diagrams of differing structure and complexity, so a
within-image comparison is more stable than an absolute baseline.

For each input image $x$ we sample a group of $G$ completions
$\{y_i\}_{i=1}^{G}\sim\pi_{\theta_{\mathrm{old}}}(\cdot\mid x)$ and score each with
the composite reward $R(y_i)$ of Eq.~\eqref{eq:reward}. We use the group mean as
the baseline and, following the reward-scaling ablation of prior
work~\cite{liu2025understanding}, do \emph{not} normalize by the group standard
deviation:
\begin{equation}
    \hat{A}_i = R(y_i) - \mu_{\mathcal{G}},
    \qquad
    \mu_{\mathcal{G}} = \frac{1}{G}\sum_{j=1}^{G} R(y_j).
\end{equation}
The update thus depends only on whether a completion is better or worse than the
others for the \emph{same} image, not on its absolute reward.

We maximize the clipped surrogate objective with a KL penalty to a frozen
reference policy $\pi_{\mathrm{ref}}$:
\begin{equation}
    \mathcal{J}_{\mathrm{GRPO}}(\theta)=
    \mathbb{E}\!\left[
    \frac{1}{G}\sum_{i=1}^{G}
    \min\!\bigl(
    \rho_i \hat{A}_i,\;
    \operatorname{clip}(\rho_i, 1-\epsilon, 1+\epsilon)\hat{A}_i
    \bigr)
    -\beta\,
    \mathbb{D}_{\mathrm{KL}}\!\left[
    \pi_\theta(\cdot\mid x)\,\|\,\pi_{\mathrm{ref}}(\cdot\mid x)
    \right]
    \right],
    \label{eq:grpo}
\end{equation}
where $\rho_i=\pi_\theta(y_i\mid x)/\pi_{\theta_{\mathrm{old}}}(y_i\mid x)$ is the
importance ratio, $\epsilon$ the clipping threshold, and $\beta$ the KL weight.
The clip bounds the per-step policy change and the KL term keeps the policy near
the reference, together stabilizing optimization of our non-differentiable,
render-gated perceptual reward.

\vspace{0.5em}\noindent\textbf{Off-policy correction.}
For throughput, rollouts are produced by a served \textsc{vLLM} snapshot of the
policy while gradients are computed by a separate training process; small
numerical differences between the two make the sampled log-probabilities a biased
proxy for $\pi_{\theta_{\mathrm{old}}}$. We correct this with token-level
truncated importance sampling: per-token ratios are clipped from above at a cap
$C$ before entering Eq.~\eqref{eq:grpo}, which bounds the variance introduced by
the generator/optimizer mismatch without discarding otherwise-valid rollouts. Throughout, we use $G=4$, $\epsilon=0.2$, $\beta=0.04$, and $C=3$.

\begin{figure}[!htbp]
    \centering
    \includegraphics[width=1\linewidth]{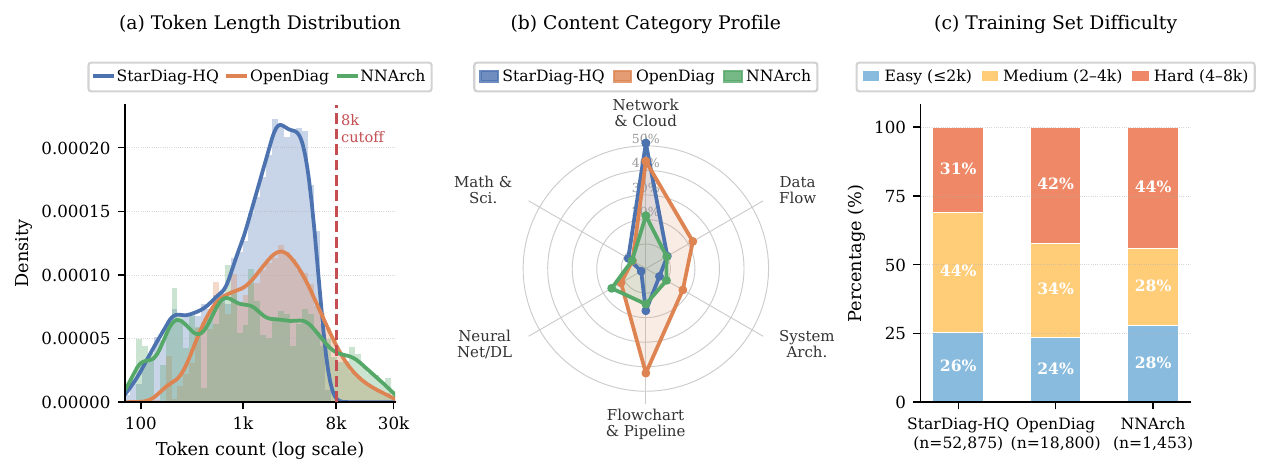}
    \caption{\textbf{StructHub dataset statistics.} Token-length distributions, content profiles, and difficulty compositions of StarDiag-HQ, OpenDiag, and NNArch.}
    \label{fig:dataset_statistics_overview}
    \vspace{-10pt}
\end{figure}

\section{Data Curation}

We introduce \textbf{StructHub}, a dataset of \textbf{84,121} structured diagram SVGs from three sources. Each sample contains a $512\times512$ rasterized PNG rendering, generated with a Chromium/Playwright renderer, paired with its ground-truth SVG code for image-to-code training and evaluation.

\textbf{StarDiag-HQ} is derived from the web-scraped StarVector dataset~\cite{rodriguez2025starvector}. We retain high-resolution SVGs with sufficient structural-text density and shape-element proportion, while removing icon-like or illustration-heavy samples. All files are capped at 8,192 tokens, resulting in \textbf{52,875} training SVGs. This subset mainly covers network/cloud architecture diagrams, flowcharts, pipelines, and data-flow graphs, providing a large and relatively clean source of structured diagrams.

\textbf{OpenDiag} is collected from open-source and open-knowledge repositories, including GitHub, Wikimedia Commons, and diagrams produced by tools such as draw.io, PlantUML, and Excalidraw. We apply a conservative SVG cleaning pipeline to remove raster images, editor metadata, layout bloat, unused definitions, empty groups, and overly long files, while normalizing code structure and converting simple paths into primitive shapes. The subset contains \textbf{27,953} SVGs, of which \textbf{18,800} satisfy the 8,192-token training threshold. Compared with StarDiag-HQ, OpenDiag contains more diverse and structurally complex real-world diagrams.

\textbf{NNArch} targets neural network and deep learning architecture diagrams, which are underrepresented in general SVG corpora. It is built from curated academic, blog, repository, and licensed diagram sources. This subset contains \textbf{3,293} SVGs, with \textbf{1,453} retained for training after token filtering. NNArch is intentionally more complex and path-heavy, capturing the dense topology of modern neural architecture figures.

\begin{wraptable}{r}{0.48\textwidth}
\vspace{-15pt}
\centering
\scriptsize
\setlength{\tabcolsep}{3pt}
\renewcommand{\arraystretch}{1.08}
\caption{Quantitative statistics of StructHub dataset sources.}
\label{tab:dataset_counts_vertical}
\resizebox{\linewidth}{!}{
\begin{tabular}{lcccc}
\toprule
\textbf{Subset}
& \makecell[c]{\textbf{Train} \\ \textbf{(Full)}}
& \makecell[c]{\textbf{Train} \\ \textbf{($\leq$8,192 tok)}}
& \makecell[c]{\textbf{Retention}}
& \makecell[c]{\textbf{Benchmark}} \\
\midrule

StarDiag-HQ
& 52,875
& 52,875
& 100.0\%
& 569 \\

OpenDiag
& 27,953
& 18,800
& 67.3\%
& 334 \\

NNArch
& 3,293
& 1,453
& 46.6\%
& 97 \\

\midrule
\textbf{Total}
& \textbf{84,121}
& \textbf{73,128}
& \textbf{86.9\%}
& \textbf{1,000} \\

\bottomrule
\end{tabular}}
\label{tab:dataset_statistics}
% \vspace{-10pt}
\end{wraptable}

\textbf{StructHub Benchmark.}
For evaluation, we build a held-out benchmark, \textbf{StructHub-Benchmark}, with \textbf{1,000 samples} from all three subsets: 569 from StarDiag-HQ, 334 from OpenDiag, and 97 from NNArch. We split the benchmark by ground-truth SVG length into three difficulty tiers: \textit{easy} ($\leq$2,048 tokens; $n=333$), \textit{medium} ($2,048$--$4,096$ tokens; $n=333$), and \textit{hard} ($>$4,096 tokens; $n=334$). This enables evaluation across different levels of structural complexity. All benchmark samples are withheld from training, with no data augmentation or overlap beyond the source-level separation.

\textbf{Dataset Statistics.}
The dataset statistics are summarized in Table~\ref{tab:dataset_statistics} and Figure~\ref{fig:dataset_statistics_overview}. 
The 8,192-token threshold affects the subsets differently: StarDiag-HQ remains unchanged, OpenDiag retains roughly two-thirds, and NNArch retains less than half, reflecting its long and topologically dense architecture diagrams. 
The excluded long-tail samples are kept to evaluate generalization beyond the training context length.

Data collection remains central but challenging, as editable structured images are manually created and hard to scale. This makes StructHub-Full valuable for structured image recovery. Representative examples are shown in Appendix Figure~\ref{fig:dataset}, and the dataset will be released to support research.

\definecolor{sectiongray}{RGB}{242,242,242}
\definecolor{methodblue}{RGB}{226,233,246}

\begin{table}[t]
\centering
\caption{
\textbf{Image-level and text-level evaluation on StructHub across difficulty levels.}
SR (render success rate) is reported as a percentage; each axis (Struct, Flow, Style) is rated on a $0$--$5$ scale and
the Overall score is scaled to $0$--$100$. 
}
\scriptsize
\setlength{\tabcolsep}{4pt}
\renewcommand{\arraystretch}{1.08}
\resizebox{0.9\linewidth}{!}{
\begin{tabular}{lcccc|cccc}
\toprule
\multirow{2}{*}{\textbf{Model}}
& \multicolumn{4}{c|}{\textbf{Image-level Evaluation}}
& \multicolumn{4}{c}{\textbf{Text-level Evaluation (GPTscore)}} \\

& \textbf{SR} $\uparrow$
& \textbf{DINO} $\uparrow$
& \textbf{LPIPS} $\downarrow$
& \textbf{SSIM} $\uparrow$
& \textbf{Overall} $\uparrow$
& \textbf{Struct} $\uparrow$
& \textbf{Flow} $\uparrow$
& \textbf{Style} $\uparrow$ \\
\midrule

\rowcolor{sectiongray}
\multicolumn{9}{c}{\textbf{\textit{Overall}}} \\
Qwen3-VL-30B-A3B
& 18.8 & 0.1566 & 0.9067 & 0.1201 & 13.29 & 0.78 & 0.68 & 0.53 \\
StarVector-8B
& 31.4 & 0.2694 & 0.7936 & 0.2170 & 21.05 & 1.04 & 1.09 & 1.03 \\
Qwen2.5-VL-7B
& \underline{52.3} & 0.3527 & 0.7834 & \underline{0.3276} & 27.56 & 1.65 & 1.32 & 1.16 \\
Qwen2.5-VL-7B$_{\mathrm{SFT}}$
& 42.8 & \underline{0.3715} & \underline{0.7403} & 0.2891 & \underline{33.49} & \underline{1.80} & \underline{1.65} & \underline{1.57} \\
\rowcolor{methodbg}
\textbf{Back2Struct (RL)}
& \textbf{71.5} & \textbf{0.5622} & \textbf{0.6473} & \textbf{0.4966} & \textbf{51.85} & \textbf{2.87} & \textbf{2.60} & \textbf{2.31} \\

\midrule
\rowcolor{sectiongray}
\multicolumn{9}{c}{\textbf{\textit{Easy}}} \\
Qwen3-VL-30B-A3B
& 26.5 & 0.2231 & 0.8659 & 0.1702 & 19.51 & 1.15 & 1.00 & 0.78 \\
StarVector-8B
& 46.1 & 0.3923 & 0.7023 & 0.3123 & 31.90 & 1.61 & 1.74 & 1.44 \\
Qwen2.5-VL-7B
& \underline{69.3} & 0.4776 & 0.7123 & \underline{0.4196} & 39.66 & 2.29 & 2.04 & 1.61 \\
Qwen2.5-VL-7B$_{\mathrm{SFT}}$
& 56.0 & \underline{0.4866} & \underline{0.6630} & 0.3789 & \underline{44.58} & \underline{2.39} & \underline{2.30} & \underline{1.99} \\
\rowcolor{methodbg}
\textbf{Back2Struct (RL)}
& \textbf{84.0} & \textbf{0.6384} & \textbf{0.5956} & \textbf{0.5796} & \textbf{62.63} & \textbf{3.44} & \textbf{3.25} & \textbf{2.70} \\

\midrule
\rowcolor{sectiongray}
\multicolumn{9}{c}{\textbf{\textit{Medium}}} \\
Qwen3-VL-30B-A3B
& 19.6 & 0.1631 & 0.9004 & 0.1288 & 13.38 & 0.78 & 0.68 & 0.54 \\
StarVector-8B
& 37.0 & 0.3232 & 0.7533 & 0.2583 & 25.27 & 1.23 & 1.25 & 1.31 \\
Qwen2.5-VL-7B
& \underline{51.8} & 0.3546 & 0.7816 & \underline{0.3306} & 26.88 & 1.62 & 1.25 & 1.16 \\
Qwen2.5-VL-7B$_{\mathrm{SFT}}$
& 45.5 & \underline{0.3985} & \underline{0.7161} & 0.3091 & \underline{36.50} & \underline{1.98} & \underline{1.76} & \underline{1.74} \\
\rowcolor{methodbg}
\textbf{Back2Struct (RL)}
& \textbf{75.6} & \textbf{0.6088} & \textbf{0.6182} & \textbf{0.5275} & \textbf{55.14} & \textbf{3.08} & \textbf{2.72} & \textbf{2.47} \\

\midrule
\rowcolor{sectiongray}
\multicolumn{9}{c}{\textbf{\textit{Hard}}} \\
Qwen3-VL-30B-A3B
& 10.2 & 0.0836 & 0.9536 & 0.0614 & 7.01 & 0.41 & 0.36 & 0.28 \\
StarVector-8B
& 11.1 & 0.0932 & 0.9249 & 0.0808 & 6.02 & 0.28 & 0.28 & 0.34 \\
Qwen2.5-VL-7B
& \underline{35.7} & 0.2263 & 0.8561 & \underline{0.2331} & 16.17 & 1.03 & 0.67 & 0.72 \\
Qwen2.5-VL-7B$_{\mathrm{SFT}}$
& 27.0 & \underline{0.2297} & \underline{0.8413} & 0.1795 & \underline{19.43} & \underline{1.04} & \underline{0.89} & \underline{0.98} \\
\rowcolor{methodbg}
\textbf{Back2Struct (RL)}
& \textbf{55.0} & \textbf{0.4399} & \textbf{0.7278} & \textbf{0.3830} & \textbf{37.83} & \textbf{2.09} & \textbf{1.83} & \textbf{1.76} \\

\bottomrule
\end{tabular}%
}
\label{tab:combined_vertical_results}
\end{table}

\begin{figure}
    \centering
    \includegraphics[width=1\linewidth]{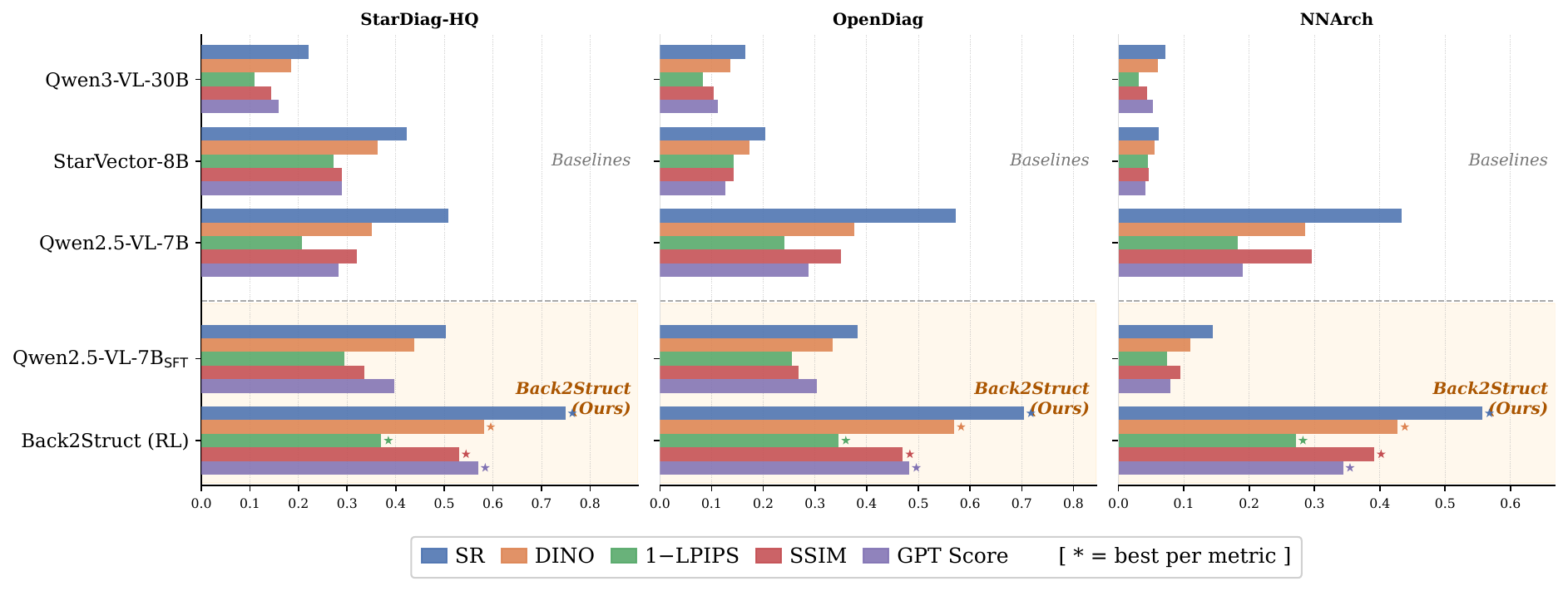}
    \caption{\textbf{Image-level performance comparison across different dataset sources.}}
    \label{fig:performance_by_sources}
    \vspace{-15pt}
\end{figure}

\section{Experiments}
\label{sec:experiment}

\subsection{Metrics}

\noindent\textbf{Image Evaluation.}
Unlike natural images, structured images (\emph{e.g.}, diagrams, charts, UI layouts) comprise discrete, symbolic elements with explicit spatial and logical relationships, whereas natural images depict continuous scenes with semantics that are implicit in dense pixel values. Consequently, instead of evaluating similarity purely at the pixel level, we assess it at the structural level.

\definecolor{sectiongray}{RGB}{242,242,242}
\definecolor{methodblue}{RGB}{226,233,246}

\begin{wraptable}{r}{0.5\textwidth}
\centering
\vspace{-1em}
\caption{
\textbf{Comparison results on cases with successfully compiled SVG / XML outputs.}
}
\scriptsize
\setlength{\tabcolsep}{2.5pt}
\renewcommand{\arraystretch}{1.1}
\begin{tabular}{lcccc}
\toprule
\textbf{Model} & DINO $\uparrow$ & LPIPS $\downarrow$ & SSIM $\uparrow$ & GPT $\uparrow$ \\
\midrule

\rowcolor{sectiongray}
\multicolumn{5}{c}{\textbf{\textit{Overall}}} \\
Qwen2.5-VL-7B      & 0.6749 & 0.5856 & 0.6270 & 52.73 \\
Qwen3-VL-30B-A3B   & 0.8347 & 0.5027 & 0.6402 & 70.88 \\
Gemini-2.5-Pro     & \textbf{0.8855} & 0.4648 & 0.6532 & \underline{77.95} \\
GPT-5              & \underline{0.8852} & \underline{0.4350} & \underline{0.6573} & 77.21 \\
\rowcolor{methodbg}
\textbf{Back2Struct} & 0.8697 & \textbf{0.3896} & \textbf{0.6738} & \textbf{78.63} \\

\midrule
\rowcolor{sectiongray}
\multicolumn{5}{c}{\textbf{\textit{Easy}}} \\
Qwen2.5-VL-7B      & 0.6894 & 0.5848 & 0.6057 & 57.25 \\
Qwen3-VL-30B-A3B   & 0.8418 & 0.4942 & 0.6421 & 73.60 \\
Gemini-2.5-Pro     & \underline{0.8818} & 0.4495 & 0.6668 & \underline{77.34} \\
GPT-5              & \textbf{0.8935} & \underline{0.4038} & \textbf{0.6718} & 77.31 \\
\rowcolor{methodbg}
\textbf{Back2Struct} & 0.8689 & \textbf{0.4012} & \underline{0.6673} & \textbf{80.32} \\

\midrule
\rowcolor{sectiongray}
\multicolumn{5}{c}{\textbf{\textit{Medium}}} \\
Qwen2.5-VL-7B      & 0.6845 & 0.5785 & 0.6381 & 51.88 \\
Qwen3-VL-30B-A3B   & 0.8333 & 0.4915 & \underline{0.6581} & 68.36 \\
Gemini-2.5-Pro     & \textbf{0.8864} & 0.4656 & 0.6471 & \underline{79.32} \\
GPT-5              & \underline{0.8840} & \underline{0.4426} & 0.6504 & 77.29 \\
\rowcolor{methodbg}
\textbf{Back2Struct} & 0.8829 & \textbf{0.3661} & \textbf{0.6807} & \textbf{79.37} \\

\midrule
\rowcolor{sectiongray}
\multicolumn{5}{c}{\textbf{\textit{Hard}}} \\
Qwen2.5-VL-7B      & 0.6332 & 0.5974 & \underline{0.6522} & 45.24 \\
Qwen3-VL-30B-A3B   & 0.8190 & 0.5458 & 0.6011 & 68.63 \\
Gemini-2.5-Pro     & \textbf{0.8883} & 0.4793 & 0.6457 & \textbf{77.10} \\
GPT-5              & \underline{0.8781} & \underline{0.4585} & 0.6498 & \underline{77.02} \\
\rowcolor{methodbg}
\textbf{Back2Struct} & 0.8461 & \textbf{0.4077} & \textbf{0.6761} & 73.20 \\

\bottomrule
\end{tabular}
\vspace{-5em}
\label{tab:llm_2x2}
\end{wraptable}

We use DINO Score~\cite{oquab2023dinov2}, LPIPS~\cite{zhang2018LPIPS}, and SSIM~\cite{wang2004SSIM} to measure the similarity between rendered SVGs and ground-truth images. We use DINO Score, LPIPS, and SSIM to measure perceptual and structural similarity~\cite{rodriguez2025starvector}, and omit MSE since our goal is faithful structural and stylistic reconstruction rather than pixel-level replication.
Additionally, we report the success rate (SR), defined as the proportion of SVG outputs that compile (render) successfully without any auxiliary SVG tools.

\noindent\textbf{Text Evaluation.}
Structured images are well captured by an object-level SVG representation. However, image-level similarity metrics may not fully reflect how effectively the model transfers semantic elements into an editable SVG format. To address this, we introduce a rubric-based GPTScore that evaluates structural similarity, flow correctness, and stylistic authenticity based on the generated SVG code. Further details are provided in the Appendix.

\definecolor{sectiongray}{RGB}{245,245,245}
\definecolor{methodbg}{RGB}{238,241,247}

\begin{wraptable}{r}{0.50\textwidth}
\vspace{-2em}
\centering
\caption{\textbf{Text-to-SVG generalization by fine-tuning on text-to-SVG task.}}
\scriptsize
\setlength{\tabcolsep}{3pt}
\renewcommand{\arraystretch}{1.05}

\begin{tabular}{lcccc}
\toprule
\textbf{Model} & DINO $\uparrow$ & LPIPS $\downarrow$ & FID $\downarrow$ & CLIP $\uparrow$ \\
\midrule

\rowcolor{sectiongray}
\multicolumn{5}{c}{\textit{Easy}} \\

Qwen3-30B & 0.8618 & \textbf{0.3218} & 77.11 & \underline{8.9622} \\
GPT-5  & 0.8294 & 0.3726 & 88.52 & 8.8018 \\
Gemini-2.5-Pro & \underline{0.8802} & \underline{0.3296} & \underline{67.72} & 8.8342 \\

\rowcolor{methodbg}
\textbf{B2S (Ours)} & \textbf{0.8987} & 0.3515 & \textbf{64.82} & \textbf{9.8271} \\

\midrule
\rowcolor{sectiongray}
\multicolumn{5}{c}{\textit{Medium}} \\

Qwen3-30B & 0.8917 & 0.3137 & 78.20 & \underline{6.6514} \\
GPT-5  & 0.8596 & 0.3577 & 92.63 & 6.5073 \\
Gemini-2.5-Pro & \underline{0.9070} & \underline{0.3114} & \underline{72.18} & 6.0003 \\

\rowcolor{methodbg}
\textbf{B2S (Ours)} & \textbf{0.9307} & \textbf{0.2877} & \textbf{44.86} & \textbf{8.1660} \\

\midrule
\rowcolor{sectiongray}
\multicolumn{5}{c}{\textit{Hard}} \\

Qwen3-30B & 0.8630 & 0.3263 & 69.92 & 6.4930 \\
GPT-5  & 0.8673 & 0.3450 & 74.73 & 5.8403 \\
Gemini-2.5-Pro & \underline{0.9058} & \textbf{0.3059} & \textbf{54.61} & \underline{6.9166} \\

\rowcolor{methodbg}
\textbf{B2S (Ours)} & \textbf{0.9310} & \underline{0.3131} & \underline{55.84} & \textbf{7.1958} \\

\bottomrule
\end{tabular}
\vspace{-5em}
\label{tab:text_2_image}
\end{wraptable}

\subsection{Experimental Setup}

We first fine-tune Qwen2.5-VL-7B~\cite{bai2025qwen25vl} with LoRA for $1$ epoch on 73K training samples, and then optimize Back2Struct with GRPO~\cite{shao2024deepseekmath_grpo} for another $3$ epoch on 4K samples from OpenDiag and NNArch. All experiments are conducted on 4 NVIDIA RTX PRO 6000 Blackwell GPUs.

For raster evaluation, we use CairoSVG~\cite{carlier2020deepsvg} for SVG rendering: instead of padding white edges, we preserve the original SVG viewBox before resizing predictions and references to $512\times512$ to avoid inflated scores from white backgrounds; we fully penalize the predictions that cannot be succesfully rendered.
We split StructHub by target SVG length into \emph{easy} ($\leq2048$ tokens), \emph{medium} ($2048$--$4096$ tokens), and \emph{hard} ($>4096$ tokens), with a maximum length of $8192$ tokens.

\subsection{Comparisons}

\noindent\textbf{StructHub Evaluation.}

Most existing Image-to-SVG methods focus on vectorizing icons, shapes, emojis, fonts, and other simple graphic assets, while our goal is to recover \textbf{editable} structured images through SVG code generation.
To the best of our knowledge, StarVector~\cite{rodriguez2025starvector} is the state-of-the-art model for diagram-oriented SVG generation, so we mainly compare Back2Struct with StarVector-8B~\cite{rodriguez2025starvector} and Qwen3-VL-8B~\cite{bai2025qwen25vl} on Image-to-SVG tasks.
Although RLRF~\cite{rodriguez2025rendering} is a more recent model, its code and weights have not been released, making direct comparison infeasible.
As shown in Table~\ref{tab:combined_vertical_results}, Back2Struct achieves the best results on DINO, LPIPS, SSIM, and GPTScore, while also producing the most reliable SVG code with the highest compile success rate.
In contrast, StarVector only generates stable SVGs when the maximum token budget is below $\sim$4000, and its performance drops sharply on the \emph{hard} split. Figure~\ref{fig:performance_by_sources} demonstrates the comparison results grouped by dataset source.

\noindent\textbf{Performance on Valid Outputs.}
For users, what matters more is the final quality of the generated results, rather than whether the model needs to regenerate a few times after failed attempts. Therefore, Table~\ref{tab:llm_2x2} summarizes the performance on examples that are successfully compiled. Our model, with only 7B parameters, outperforms commercial models such as GPT-5~\cite{singh2025openai_gpt5} and Gemini-2.5-Pro~\cite{comanici2025gemini25}. Since querying these models is very expensive, we randomly sampled 200 samples from the StructHub dataset for this comparison.

\noindent\textbf{Text-to-SVG Generalization.} 
Our method can also support text-to-SVG generation. We further fine-tune the image-to-SVG Back2Struct model on a text-to-SVG dataset and summarize the results in Table~\ref{tab:text_2_image}. Promisingly, our method remains competitive with GPT-5~\cite{singh2025openai_gpt5} and Gemini-2.5-Pro~\cite{comanici2025gemini25}, and even shows clear advantages.

\noindent\textbf{Qualitative Evaluation.}
Figure~\ref{fig:first_show} compares \model against GPT-5-Thinking~\cite{achiam2023gpt4}, Qwen3-MAX~\cite{qwen3}, and Gemini-2.5-Pro~\cite{comanici2025gemini25} on Image-to-SVG and Text-to-SVG generation. Where these ba
selines miss shapes, misalign arrows, or misplace text, \model{} better preserves fine-grained spatial layout, object hierarchy, and inter-element relations, producing structurally and stylistically faithful diagrams---
consistent with our quantitative results.

\begin{table}[t]
\centering
\caption{
\textbf{Ablation on reward design across difficulty tiers.}
S/F/P denote the syntactic, fidelity, and perceptual rewards; R/L/-- denote
rule-based, LLM-based, and unused. The active rewards are combined either by
summation ($\Sigma$) or by our render-gated composition (\emph{gate}).
}
\scriptsize
\setlength{\tabcolsep}{1.45pt}
\renewcommand{\arraystretch}{1.2}
\begin{tabular*}{\linewidth}{@{\extracolsep{\fill}} l ccc cccc cccc cccc @{}}
\toprule
\multirow{2}{*}{\textbf{Method}}
& \multicolumn{3}{c}{\textbf{Reward}}
& \multicolumn{4}{c}{\textbf{Easy}}
& \multicolumn{4}{c}{\textbf{Medium}}
& \multicolumn{4}{c}{\textbf{Hard}} \\
\cmidrule(lr){2-4}\cmidrule(lr){5-8}\cmidrule(lr){9-12}\cmidrule(lr){13-16}
& S & F & P
& SR $\uparrow$  & DINO  $\uparrow$ & LPIPS $\downarrow$ & SSIM $\uparrow$ 
& SR $\uparrow$  & DINO  $\uparrow$ & LPIPS $\downarrow$ & SSIM $\uparrow$ 
& SR $\uparrow$  & DINO  $\uparrow$ & LPIPS $\downarrow$ & SSIM $\uparrow$  \\
\midrule

Base\,+\,SFT & -- & -- & --
& 56.0 & 0.4866 & 0.6630 & 0.3789
& 45.5 & 0.3985 & 0.7161 & 0.3091
& 27.0 & 0.2297 & 0.8413 & 0.1795 \\

\multirow{4}{*}{\scalebox{1.4}{$\Downarrow$}$+$RL$_{\Sigma}$} & -- & -- & L
& 56.3 & 0.4898 & \underline{0.6535} & 0.3890
& 45.5 & 0.3965 & 0.7185 & 0.3051
& 23.1 & 0.1985 & 0.8625 & 0.1551 \\

 & R & -- & L
& 55.7 & 0.4797 & 0.6597 & 0.3827
& 46.4 & 0.4070 & 0.7111 & 0.3125
& 25.2 & 0.2143 & 0.8536 & 0.1649 \\

 & R & R & L
& 55.7 & 0.4842 & 0.6663 & 0.3718
& 45.8 & 0.4042 & \underline{0.7098} & 0.3117
& 23.4 & 0.1982 & 0.8613 & 0.1584 \\

 & R & R & R
& \underline{74.4} & \underline{0.5184} & 0.6904 & \underline{0.4434}
& \underline{59.3} & \underline{0.4072} & 0.7589 & \underline{0.3713}
& \underline{40.5} & \underline{0.2684} & \underline{0.8387} & \underline{0.2594} \\

\cmidrule(lr){1-16}
\rowcolor{methodbg}
$\rightarrow$RL$_{\mathrm{gate}}$ & R & -- & R
& \textbf{84.0} & \textbf{0.6384} & \textbf{0.5956} & \textbf{0.5796}
& \textbf{75.6} & \textbf{0.6088} & \textbf{0.6182} & \textbf{0.5275}
& \textbf{55.0} & \textbf{0.4399} & \textbf{0.7278} & \textbf{0.3830} \\

{\scriptsize\itshape\textcolor{goodgreen}{$\Delta$}} &  &  & 
& {\tiny\itshape\textcolor{goodgreen}{(+28.0)}} & {\tiny\itshape\textcolor{goodgreen}{(+.152)}} & {\tiny\itshape\textcolor{goodgreen}{(+.067)}} & {\tiny\itshape\textcolor{goodgreen}{(+.201)}}
& {\tiny\itshape\textcolor{goodgreen}{(+30.1)}} & {\tiny\itshape\textcolor{goodgreen}{(+.210)}} & {\tiny\itshape\textcolor{goodgreen}{(+.098)}} & {\tiny\itshape\textcolor{goodgreen}{(+.218)}}
& {\tiny\itshape\textcolor{goodgreen}{(+28.0)}} & {\tiny\itshape\textcolor{goodgreen}{(+.210)}} & {\tiny\itshape\textcolor{goodgreen}{(+.114)}} & {\tiny\itshape\textcolor{goodgreen}{(+.204)}} \\

\bottomrule
\end{tabular*}\vspace{-2em}
\label{tab:ablation}
\end{table}

\subsection{Ablation Study}
\label{sec:abla}

Table~\ref{tab:ablation} ablates our reward along two axes: signal choice and
composition. From the SFT baseline, RL with only partial or LLM-based signals
$(\times,\text{L})$, $(\text{R},\text{L})$, or $(\text{L},\text{L})$ yields small,
unstable gains: Easy render success is flat ($56.0\!\to\!55.7$--$56.3$) and Hard
worsens ($27.0\!\to\!23.1$--$25.2$), indicating that a single/noisy signal cannot
reliably guide SVG recovery. Using both rule-based signals with a \emph{summed}
reward ($\Sigma$, R/R) improves SR (Easy $56.0\!\to\!74.4$, Hard $27.0\!\to\!40.5$)
but degrades perceptual quality, increasing LPIPS on Easy/Medium
($0.6630\!\to\!0.6904$, $0.7161\!\to\!0.7589$).

The main driver is \emph{composition}. Replacing summation with our render-gated
composition—keeping the same rule-based components and training data—improves all
metrics. Gating boosts SR over the summed reward by $+9.6/+16.3/+14.5$ on
Easy/Medium/Hard ($74.4\!\to\!84.0$, $59.3\!\to\!75.6$, $40.5\!\to\!55.0$) and
reduces LPIPS at every tier ($0.6904\!\to\!0.5956$, $0.7589\!\to\!0.6182$,
$0.8387\!\to\!0.7278$), eliminating the summation trade-off. Relative to SFT, the
gated reward increases SR by $+28.0/+30.1/+28.0$, improves DINO by up to $+0.21$
and SSIM by up to $+0.22$, and reduces LPIPS by $0.067/0.098/0.114$. Since summed
and gated variants share components and data, the gap isolates composition:
gating makes syntactic and perceptual signals cooperate by treating rendering as
a prerequisite, removing the need for a separate length-fidelity reward.

\vspace{-1em}

\section{Conclusion}

\vspace{-0.5em}

In this work, we presented \model, a code-centric framework that makes structured images editable again by recovering semantically meaningful SVG / XML code from raster inputs. Rather than treating diagrams, charts, and flowcharts as ordinary images, \model represents them as symbolic structures composed of text, shapes, topology, and layout, enabling direct rendering, import into authoring tools, and downstream editing. We further introduced reward-based optimization to align training with deployment needs, jointly encouraging syntactic validity, structural fidelity, and perceptual faithfulness. Experiments on StructHub show that \model outperforms strong vision--language, code-generation, and diagram-oriented SVG baselines across image-level metrics, text-level GPTScore, compile success rate, and qualitative results. Ablation studies further confirm that the proposed reward components are complementary and necessary for reliable editable SVG recovery.

\bibliographystyle{plainnat}
\bibliography{main}

\newpage
\appendix

\begin{center}
\Large \textbf{Back2Struct: Making Structured Images Editable Again} \par 

\vspace{1em}
{\large \textbf{Supplementary}}
\end{center}

\begin{figure}[ht]
    \centering
    \includegraphics[width=1\linewidth, height=0.7\linewidth]{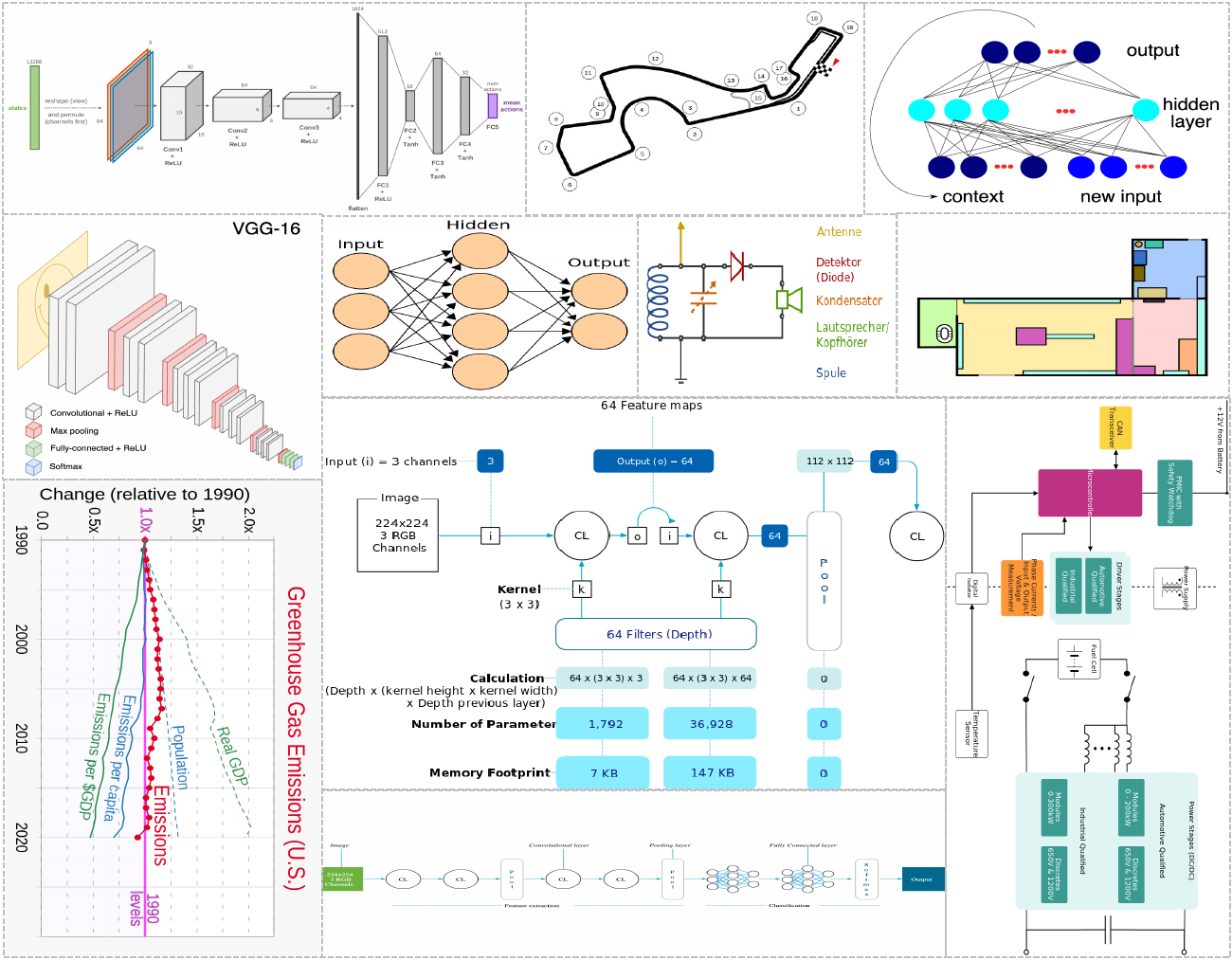}
    % \vspace{-5pt}
    \caption{\textbf{Examples from our dataset.} Each raster image is paired with an editable SVG counterpart, providing high-quality pixel--code supervision for image-to-SVG generation.}
    \label{fig:dataset}
    \vspace{-12pt}
\end{figure}

\section{StructHub Dataset}
\label{app:dataset}

\subsection{SVG Cleaning and Normalization Details}

Beyond the curating and preprocessing pipeline described for OpenDiag in the main text, all three subsets share a common post-processing step applied after source-specific cleaning: (1) the root \texttt{<svg>} element is given an explicit \texttt{width}, \texttt{height}, and \texttt{viewBox} if any are missing, ensuring unambiguous coordinate semantics for the renderer; (2) namespace prefixes (\texttt{svg:}, \texttt{xlink:}) are normalized to unprefixed equivalents wherever SVG~1.1 permits; and (3) a final UTF-8 re-encoding pass removes null bytes and non-printable control characters that occasionally appear in crawled files.
These steps are applied uniformly so that the model sees a consistent SVG dialect regardless of source.

\subsection{Benchmark Construction and Stratification}

The 1,000-sample StructHub-Benchmark is drawn from the held-out portion of each source using stratified sampling to ensure balanced representation across both source and difficulty tier.
Source proportions roughly mirror the training distribution (StarDiag-HQ: 56.9\%, OpenDiag: 33.4\%, NNArch: 9.7\%), and each source is further stratified into three equal-sized difficulty buckets ($n \approx 333$ each) based on ground-truth token count: easy ($\leq$2,048 tokens), medium (2,049--4,096 tokens), and hard ($>$4,096 tokens).
Stratification is done independently per source to avoid over-representing short StarDiag-HQ samples in the easy tier.
No sample appearing in the benchmark was ever used for training or reward computation.

\subsection{Content Diversity}

Figure~\ref{fig:dataset} shows representative examples drawn from all three subsets, illustrating the breadth of diagram types present in StructHub.
The collection spans a wide spectrum of structural diagram genres: deep neural network architectures with multi-layer topology (top-left, top-right), racetrack and schematic diagrams (top-center), electrical circuit schematics (center), floor plans and spatial layouts (center-right), data-flow and parameter-count visualizations (center), and time-series charts with annotated legends (bottom-left).
This diversity is deliberate: diagrams that share the same surface rendering (e.g., a flowchart and a circuit schematic both use boxes and arrows) may require structurally very different SVG encodings, stressing both the syntactic and semantic axes of the generation task.
The NNArch subset in particular contributes diagrams with dense path geometry and sparse text annotation---a regime where pixel-level similarity metrics diverge most sharply from structural correctness, motivating the multi-axis reward design described in Method Section.

\subsection{Licensing and Data Availability}

StarDiag-HQ is derived from the StarVector-diagram dataset, which aggregates SVGs from open web sources; we apply no additional license restrictions beyond those of the original corpus.
OpenDiag is assembled from GitHub repositories and Wikimedia Commons content released under permissive open-source licenses (MIT, Apache-2.0, CC-BY variants); repository-level license files were checked and repositories with non-permissive or proprietary licenses were excluded.
NNArch contains a subset of commercially licensed diagram assets; these samples are provided for research use only and must not be redistributed in commercial products.
We will release the full dataset (code, cleaned SVGs, rendered PNGs, and benchmark splits) upon acceptance, subject to the above license constraints.

\section{Evaluation Metrics}
In this section, we provide additional details on the evaluation setup used in our experiments and explain the rationale for adopting a protocol that differs from prior work. We further describe our text-level evaluation rubric (GPTScore), which we use to assess the quality of the generated SVG code.

\subsection{Image-level evaluation}
In the context of image-level comparison, model predictions are evaluated utilizing three prevalent similarity metrics: DinoScore, LPIPS, and SSIM. Given that LPIPS is a metric of perceptual distance where a lower value indicates better similarity, it is transformed into a similarity score via
the expression \(1-\text{LPIPS}\), ensuring
 that all evaluation metrics adhere to the ``higher-is-bette'' standard. While these metrics are conventional for assessing raster images, the evaluation of our tasks presents additional complexities: the model outputs are in the form of SVG code, thus rendering setup and management of failure cases are pivotal and can substantially influence the final scores.

\begin{figure*}[ht]
    \centering
    \includegraphics[width=\linewidth]{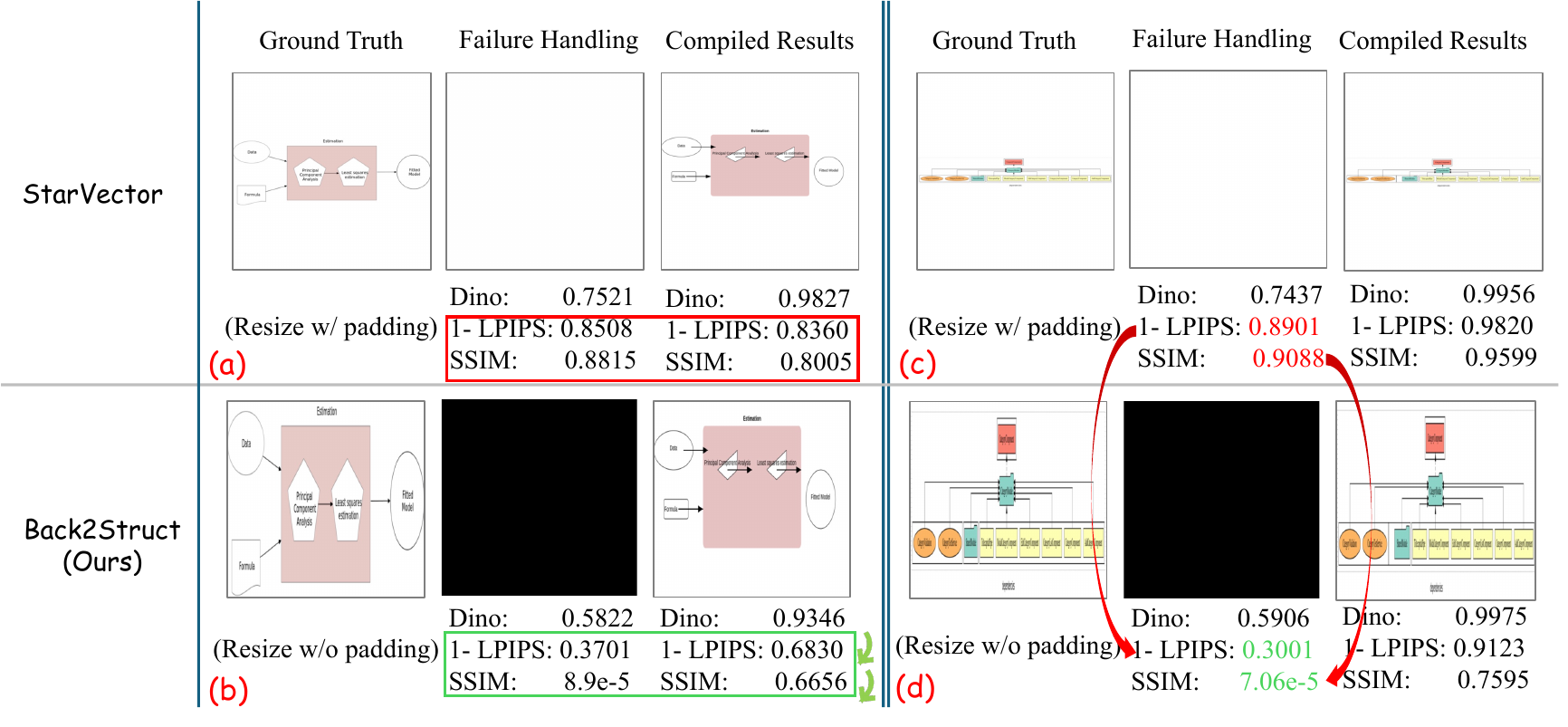}
    \caption{\textbf{An ablative comparison on image-level evaluation.} Different evaluation configurations can significantly affect the final scores, and our setup is designed to reduce background noise and provide a more reliable reflection of model performance.}
    \label{fig:imglvl_evalissue}
\end{figure*}

The rendering of SVG files involves flexibility in resolution and aspect ratio of the viewbox, where parameters such as canvas size, padding strategy, and scaling directly affect the ultimate rasterized image. Furthermore, the management of invalid or incomplete SVG predictions (e.g., SVGs that fail to compile) introduces additional variability. If not meticulously addressed, these details may obscure the actual visual discrepancies between predictions and ground-truth images. Figure~\ref{fig:imglvl_evalissue} elucidates these issues through two representative examples: one imperfect prediction (left) and one near-perfect prediction (right). For each example, we present: 1) the ground-truth image; 2) the failure-case image; 3) the rendered output of the model when the SVG compiles successfully.

Previous research, specifically StarVector, approaches rendering by displaying each SVG on a fixed square canvas, substituting failure cases with a pure white image. Conversely, our setup involves rendering SVGs while maintaining their original aspect ratio, subsequently resizing them uniformly to $512\times512$, and representing failure cases with a black image.

Under StarVector's configuration, the substantial white space introduces significant noise in the final score, as depicted in Figure~\ref{fig:imglvl_evalissue} (a), where the failure case (white image) achieves higher SSIM and \(1-\text{LPIPS}\) scores than the successfully compiled result. This problem is exacerbated in cases with higher image ratios, as demonstrated in Figure~\ref{fig:imglvl_evalissue} (c), where the failure case yields an SSIM score of 0.9088, which is unrealistically high. In our configuration, substituting the white image with a black image as the failure case markedly decreases all evaluation scores. To further mitigate noise from the background, we resize the original image instead of employing a ``padding'' operation. As a result, the evaluation score significantly declines from Figure~\ref{fig:imglvl_evalissue} (a) to Figure~\ref{fig:imglvl_evalissue} (b) on compiled but imperfect outcomes, thereby more accurately reflecting the method's performance.

\begin{figure*}[t]
    \centering
    \includegraphics[width=0.95\linewidth]{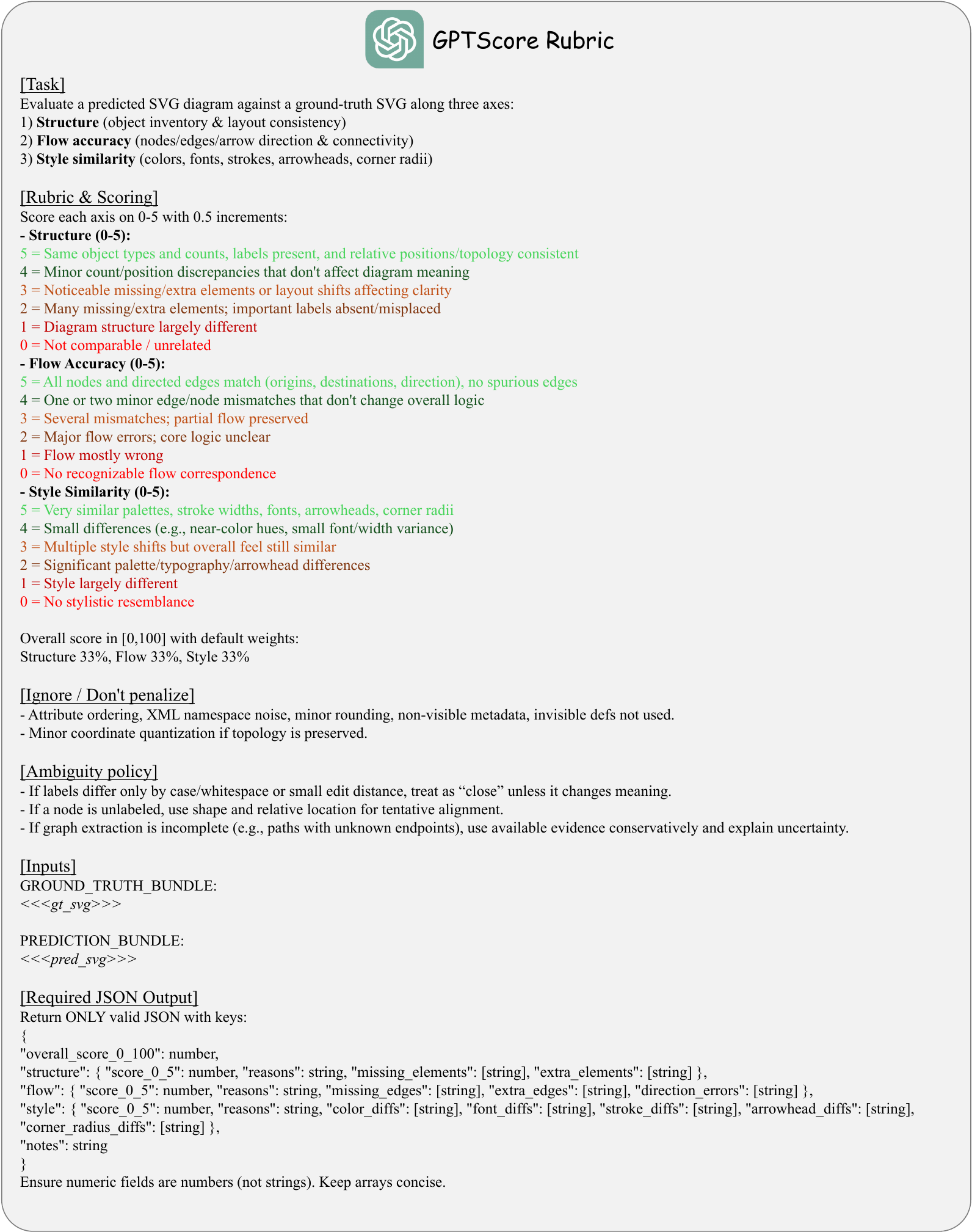}
    \caption{\textbf{Our GPTScore rubric} evaluates predicted SVG code against ground-truth SVGs along three dimensions—\textbf{Structure}, \textbf{Flow Accuracy}, and \textbf{Style Similarity}—providing a unified measure of content fidelity and visual consistency.}\vspace{-1em}
    \label{fig:gptscore_rubric}
\end{figure*}

\subsection{Text-level evaluation -- GPTScore}
In addition to metrics at the image level, we implement a text-level evaluation protocol, termed GPTScore, to appraise the structural, logical, and stylistic fidelity of the generated SVG code in comparison to the ground-truth SVG. While image-level metrics deliver a broad similarity signal predicated on rendered appearance, the text-level evaluation meticulously examines the SVG code structure itself, providing more detailed insights into whether the model accurately encapsulates objects, their interrelationships, and stylistic attributes. This is particularly crucial in our context, where the ultimate objective is to generate diagrams that are both editable and semantically accurate.

GPTScore assesses each prediction along three dimensions (as shown in Figure~\ref{fig:gptscore_rubric}): (1) Structural Integrity, (2) Flow Accuracy, and (3) Style Similarity. Each dimension is rated on a scale from 0 to 5 in 0.5-point increments, with the final score being consolidated on a 0 to 100 scale using equal weighting unless specified otherwise. Below, we delineate the design principles, scoring rubric, and interpretive guidelines employed in our evaluation.

\subsubsection{Structure Evaluation}
The \textbf{Structure} axis measures how well the predicted SVG reproduces the fundamental building blocks of the diagram, including:
(i) the inventory of objects (\textit{e.g.}, rectangles, ovals, diamonds, arrows, text labels),
(ii) their relative positions, and
(iii) the overall spatial topology.
A score of $5$ indicates that the prediction closely matches the ground truth in terms of object types, counts, labels, and layout.
Minor deviations that do not alter the semantic meaning of the diagram (\textit{e.g.}, small coordinate shifts) are tolerated in the $4$--$5$ range.
Scores of $2$--$3$ correspond to noticeable missing or extra elements and layout shifts that begin to impact clarity, while a score of $1$ indicates that the diagram structure is largely different.
A score of $0$ is assigned when the prediction is not comparable or is essentially unrelated to the reference SVG.

To avoid penalizing irrelevant differences, we instruct the evaluator to ignore attribute ordering, XML namespace noise, minor rounding, non-visible metadata, and unused \texttt{<defs>} elements.
Minor coordinate quantization is also not penalized as long as the relative topology and alignment of objects are preserved.

\subsubsection{Flow Accuracy Evaluation}
The \textbf{Flow Accuracy} axis evaluates whether the predicted SVG correctly captures the directed relationships between nodes, such as arrows in flowcharts or process diagrams.
Concretely, this includes checking:
(i) whether nodes and edges in the prediction can be aligned to those in the ground truth,
(ii) whether the origins and destinations of edges match, and
(iii) whether the arrow directions are correct and no spurious edges are introduced.

A score of $5$ indicates that all nodes and directed edges match the ground truth, with no extra or missing connections and no direction errors.
A score of $4$ allows one or two minor mismatches that do not change the overall logical flow.
Scores around $3$ indicate several mismatches where the high-level logic is still partially preserved.
Scores of $1$--$2$ correspond to major flow errors where the core logic becomes unclear or largely incorrect.
A score of $0$ is assigned when there is no recognizable flow correspondence between the prediction and the ground truth.

\begin{wrapfigure}{r}{0.48\textwidth}
    \centering
    \includegraphics[width=\linewidth]{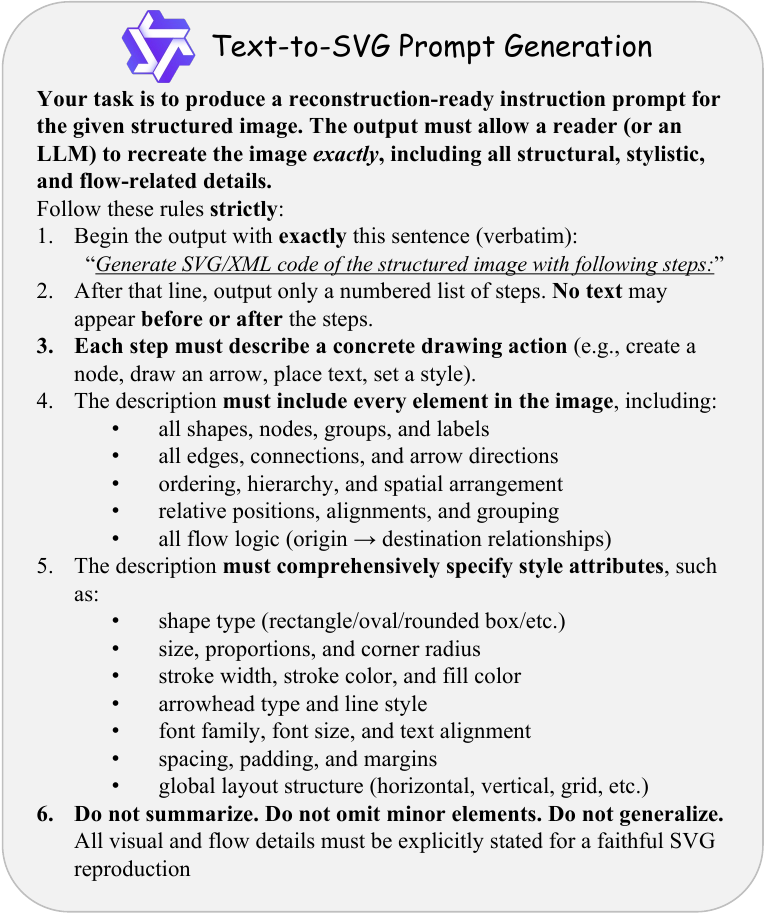}
    \caption{\textbf{Text-to-SVG instruction–prompt pair generation pipeline.} We construct paired instruction prompts and SVG outputs that encode deterministic, fine-grained drawing specifications to supervise structured image reconstruction.}\vspace{-5em}
    \label{fig:text2svg_prompt_gen}
\end{wrapfigure}

When labels differ only by case, whitespace, or small edit distance, we treat them as close matches unless the change alters the semantic meaning.
For unlabeled nodes, evaluators are instructed to use shape and relative position to tentatively align elements.
If the graph connectivity is partially ambiguous (\textit{e.g.}, paths with unclear endpoints), the evaluator uses the available evidence conservatively and notes any uncertainty in the evaluation.

\subsubsection{Style Similarity Evaluation}
The \textbf{Style Similarity} axis measures how well the predicted SVG preserves the visual appearance of the reference diagram, including:
(i) color palettes,
(ii) stroke widths,
(iii) fonts and text sizes,
(iv) arrowhead styles, and
(v) corner radii of shapes.

A score of $5$ corresponds to very close stylistic agreement across these properties such that the predicted diagram appears almost indistinguishable in style from the ground truth.
A score of $4$ indicates only small differences (\textit{e.g.} near-by color hues, slightly different fonts, or stroke widths) that do not substantially change the overall visual impression.
Scores around $3$ correspond to multiple style changes while maintaining a generally similar look and feel.
Scores of $1$--$2$ reflect significant differences in palette, typography, arrowheads, or shape geometry, resulting in a noticeably different style.
A score of $0$ is reserved for predictions that essentially do not resemble the reference stylistic.

As with the structure axis, non-visible SVG elements and unused definitions are ignored to prevent them from affecting stylistic scoring.
The focus is placed on properties that impact the rendered, human-visible appearance of the diagram.

\subsection{Creating Text-to-SVG Instruction Prompt}
We employ the Qwen2.5-VL model for the conversion of structured images into descriptions suitable for reconstruction. These descriptions serve as deterministic, human-interpretable programs that delineate precise instructions on what elements to draw and the methodology of their depiction. The instructions provide a comprehensive enumeration of objects, textual labels, shapes, edges, and arrow directions, while concurrently capturing essential stylistic attributes such as stroke widths, colors, corner radii, fonts, spacing, and overarching layout patterns. This intermediary phase significantly mitigates ambiguity and ensures that downstream Scalable Vector Graphics (SVG) generation models receive comprehensive and precise visual specifications.

To ensure consistency and reproducibility, we implement a rigorously enforced prompting format specifically designed for Large Vision Language Models (LVLMs). The model is required to produce (1) a standardized introductory phrase, (2) a pure sequence of explicit drawing steps, and (3) a thorough documentation of structural and stylistic details devoid of extraneous commentary. This controlled structure averts over-generalization, excludes irrelevant text, and provides a practicable framework for precise SVG reconstruction. Consequently, the LVLM functions not merely as a captioning system but as an exact visual-to-instruction compiler.

Figure~\ref{fig:text2svg_prompt_gen} shows detailed design principles, enforcement protocols, and the rationale underlying our instruction-prompting strategy. We demonstrate that converting structured images into explicit, step-by-step drawing instructions significantly enhances reconstruction fidelity, diminishes model hallucination, and enables the downstream LLM to produce SVG code indistinguishable in appearance from the original input image.

\begin{wrapfigure}{r}{0.48\textwidth}
    \centering
    \includegraphics[width=\linewidth]{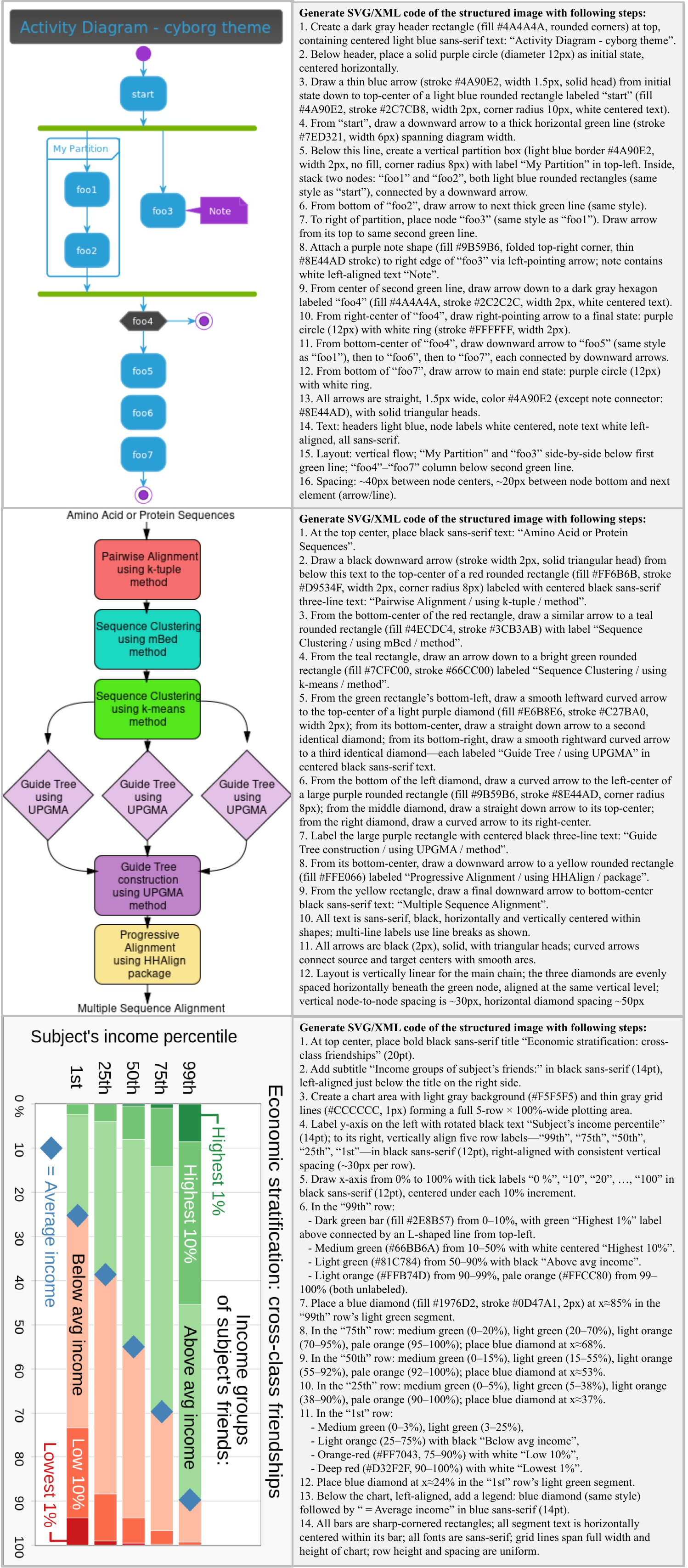}
    \caption{Example instruction prompts generated by the LVLM for guiding SVG reconstruction from structured images.}
    \label{fig:text2svg_samples}
    \vspace{-3em}
\end{wrapfigure}

Figure~\ref{fig:text2svg_samples} showcases a diverse set of Text-to-SVG instruction–image pairs, exemplifying the comprehensive scope and precision inherent in our dataset. The displayed examples encompass various structured-image domains, including an activity diagram characterized by partitions and a multi-stage control flow, a bioinformatics pipeline incorporating branching guide-tree operations, and a quantitative chart that represents socioeconomic stratification across income percentiles. These samples elucidate how our instruction prompts meticulously encode not only object types, shapes, and hierarchical arrangements, but also intricate stylistic attributes such as color palettes, stroke widths, arrow geometries, curved connectors, grid structures, and multi-line text formatting. The examples underscore the robustness of our prompting protocol in accurately capturing semantic structure alongside detailed visual style, thereby ensuring that subsequent models receive comprehensive and unambiguous specifications. This capability facilitates the reconstruction of SVGs that closely replicate the original diagrams.

\section{Qualitative Samples}
\begin{figure*}[t]
    \centering
    \includegraphics[width=0.8\linewidth]{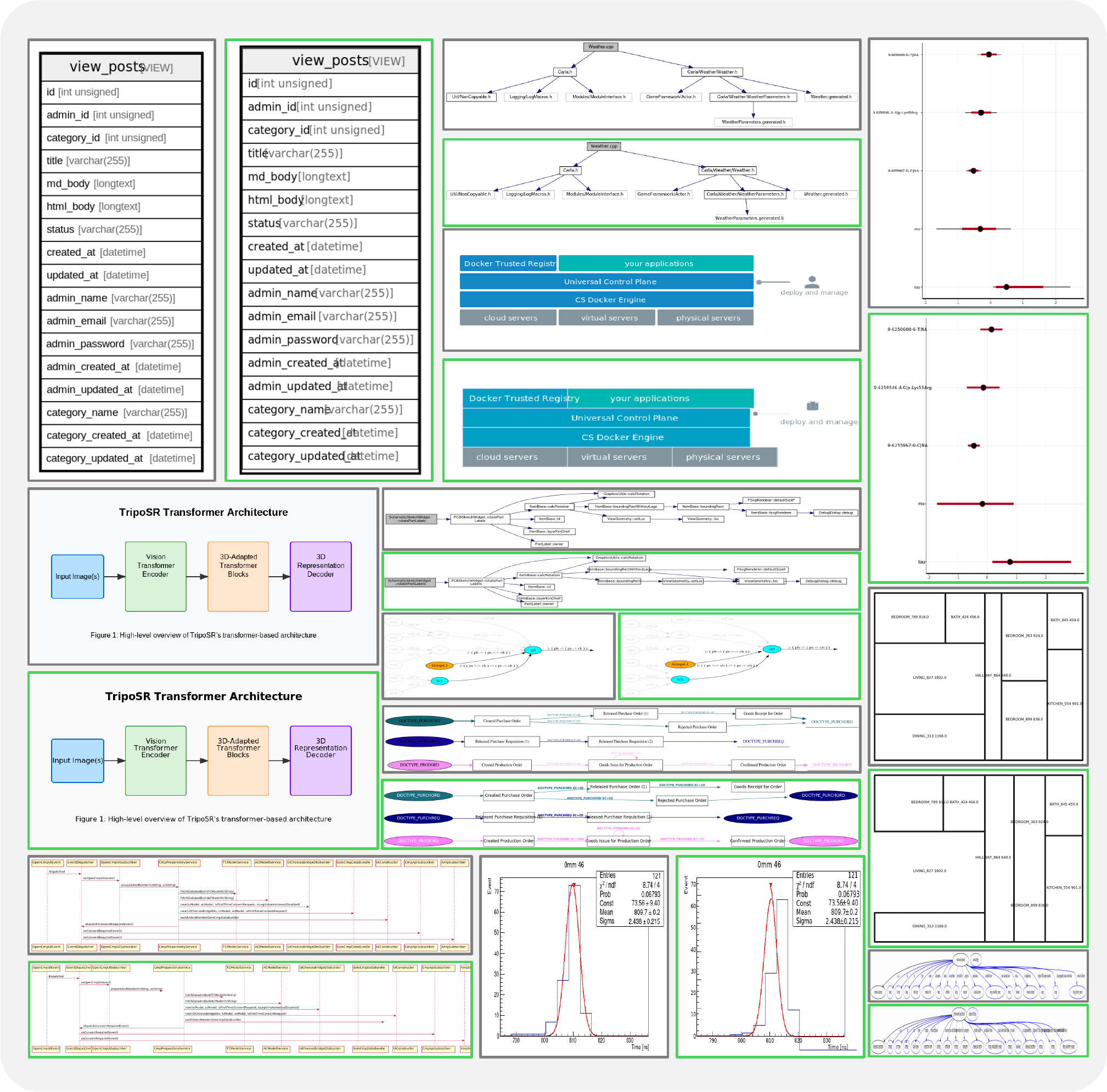}
    \caption{\textbf{Qualitative results of our method.} Gray denotes the ground-truth SVGs, while green indicates the predicted outputs.
}
    \label{fig:qualitative_sample}
\end{figure*}

Figure~\ref{fig:qualitative_sample} provides additional qualitative samples, underscoring the scope, intricacy, and robustness of our pipeline. These samples encompasse a comprehensive array of structured image types, including relational database schemas with nested attribute tables, transformer architecture diagrams, multi-stage system workflow charts, Docker deployment diagrams, statistical plots with annotated distributions, hierarchical tree structures, and block-based layout schematics. Our method adeptly captures not only the overarching semantic organization—such as table hierarchies, model pipelines, and system components—but also the nuanced geometric and stylistic elements, including alignment, spacing, connectors, arrow types, fonts, and color schemes. These qualitative outcomes demonstrate that our methodology effectively generalizes across diverse diagram styles, maintains structural fidelity across varying visual formats, and generates precise, editable SVG outputs even for densely populated or domain-specific technical graphics.

\begin{figure*}[t]
    \centering
    \includegraphics[width=0.8\linewidth, height=0.35\linewidth]{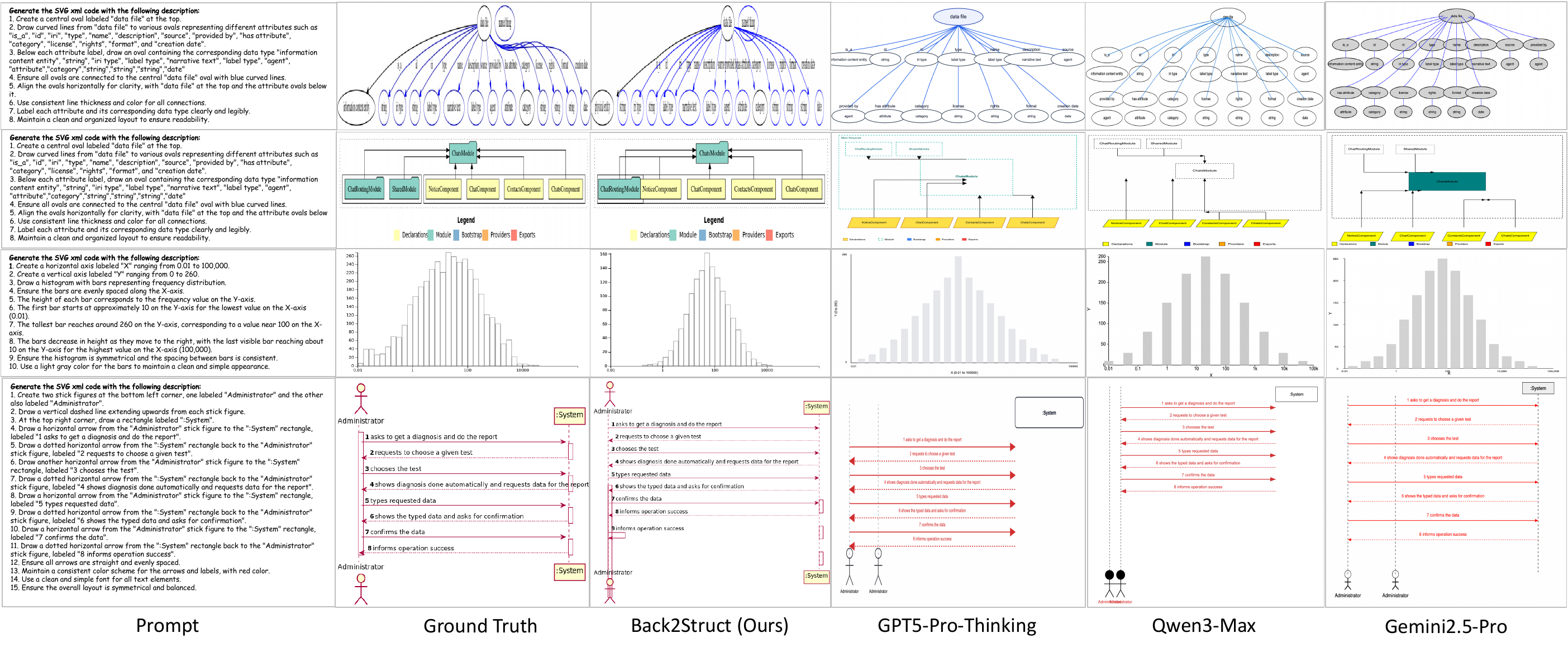}
    \caption{\textbf{A qualitative comparison on Text-to-SVG task among advanced LLMs and \model.} Given the same prompt, \model, despite being only 7B, produces structured images that are more similar to the ground truth. }
    \label{fig:comparison_prompt2svg}
    \vspace{-10pt}
\end{figure*}

Figure~\ref{fig:comparison_prompt2svg} extends this analysis to the Text-to-SVG setting, where each model must render a structured diagram directly from the same natural-language specification. Despite having only 7B parameters, \model produces SVGs that are markedly closer to the ground truth than those of substantially larger systems: it faithfully realizes the specified components, their spatial arrangement, and the connectors and labels that encode their relationships. The baseline generations, by contrast, frequently drop or duplicate elements, misplace text, or break the intended alignment and arrow topology, yielding diagrams that are visually plausible yet structurally inconsistent with the prompt. This indicates that \model{}'s reward-driven fine-tuning transfers from image-conditioned reconstruction to purely text-conditioned generation, preserving fine-grained layout and stylistic fidelity even under the weaker grounding of a textual prompt.

\section{Limitations}
\begin{itemize}
    \item Since there is no open-source code generation large language model that accepts images as input, Back2Struct is built on a non-specialized code generation model, which limits its capability. 
    \item In addition, to ensure stable training, the current model supports at most 8K output tokens, which restricts Back2Struct from generating more complex images.
\end{itemize}
We note that these limitations do not undermine the innovation and novelty of our work, and they can be overcome in future research.

\end{document}